\documentclass{article}

\usepackage[preprint, nonatbib]{neurips_2021}

\usepackage[utf8]{inputenc} 
\usepackage[T1]{fontenc}    
\usepackage{hyperref}       
\usepackage{url}            
\usepackage{booktabs}       
\usepackage{amsfonts}       
\usepackage{nicefrac}       
\usepackage{microtype}      
\usepackage{xcolor}         
\usepackage{amsmath}
\usepackage{graphicx}
\usepackage{subfig}
\graphicspath{ {./images/} }
\title{Conditional Autoregressors are Interpretable Classifiers}

\author{%
  Nathan Elazar\\
  Australian National University\\
  \texttt{nathan.elazar@anu.edu.au} \\
}

\begin{document}

\maketitle

\begin{abstract}
  We explore the use of class-conditional autoregressive (CA) models to perform image classification on MNIST-10. Autoregressive models assign probability to an entire input by combining probabilities from each individual feature; hence classification decisions made by a CA can be readily decomposed into contributions from each each input feature. That is to say, CA are inherently locally interpretable. Our experiments show that naively training a CA achieves much worse accuracy compared to a standard classifier, however this is due to over-fitting and not a lack of expressive power. Using knowledge distillation from a standard classifier, a student CA can be trained to match the performance of the teacher while still being interpretable.
\end{abstract}

\section{Introduction}

A standard classification model defines, for any given input $x\in \mathbb{R}^n$, a probability distribution over classes $P(y|x)$. Bayes rule tells us that this probability can equivalently be expressed as
$$P(y|x) \propto P(x|y)P(y)$$
That is, we can equivalently perform classification by training, for each class $y$, a conditional generative model of $P(x|y)$, and at inference time selecting the class whose generator assigns the highest probability to $x$ (weighted by class priors). If the generative model has an autoregressive structure, then we can further decompose the probability as 
$$P(y|x) \propto P(y)\prod_{i=1}^n P(x_i|x_{:i}, y)$$
Where $P(x_i|x_{:i})$ denotes the probability that feature $i$ has value $x_i$ given all previous features up to but excluding $i$, for some predefined ordering of features. In this formulation, the contribution of the $i$'th feature towards the classification decision is readily given by $P(x_i|x_{:i},y)$, which represents the amount of novel information that feature $i$ contains about class $y$. We argue that this is, by definition, the importance of feature $i$ to the model's decision.

\section{Related Work}
\subsection{Local Interpretability}
The black-box nature of machine learning models has been identified as one of the major issues limiting their wide-spread adoption. In light of this, The research community has developed numerous different approaches to help explain how models function. The most popular methods for explaining image classifiers are post-hoc feature-importance based local explanations \cite{xai_survey}. These methods, such as LIME\cite{lime}, SHAP\cite{shap}, GRAD-CAM\cite{grad_cam}, or integrated gradients\cite{integrated_gradients}, explain why a model made a particular decision for a particular input example. They do this by assigning each input feature a score, representing how important that feature's value is to the model's decision. While the development of post-hoc methods is an active area of research with numerous variations and novel methods being proposed regularly, all of these approaches have the fundamental limitation that they are \textit{surrogates} and are therefore not guaranteed to be representative of the original model \cite{stopexplain}. Furthermore, there does not exist an objective criteria by which to compare them.

\subsection{Autoregressive Models}
An autoregressive model is a generative model which decomposes the joint probability distribution over features using the chain rule of probability\cite{pixelcnn}.
$$P(x)=\prod_{i=1}^n P(x_i|x_{:i})$$
In practice, this means that input features are shown to the model one at a time, and at each step the model predicts the value of the next feature, conditioned on the values of all features seen so far. For this purpose it is necessary to decide on a (fixed) ordering to present the features in. After an autoregressor has been trained, it can be used to sample new inputs by feeding its own outputs back in as input to itself, thus building up a complete sample one feature at a time.

For the case of images specifically, numerous autoregressive architectures based on causal convolutions have been proposed \cite{pixelcnn}. These architectures are appealing because during training they are able to perform all of the independent steps of the prediction at the same time, in one forward pass of the model. This means that they can be trained in roughly the same time as a standard CNN classifier. Causal convolutions are only valid for the canonical raster ordering of pixels (left to right along each row, rows ordered top to bottom), and so we use this ordering throughout.

\subsection{Conditional Generative Models}
While a generative model gives $p(x)$, a conditional model gives $p(x|y)$, where $y$ can be any auxiliary information. Conditional models can be created by feeding in $y$ as an additional input to a generative model, or if $y$ is discrete by training an independent generative model for each value of $y$. While CA have been studied extensively in the context of conditional generation, there has been little-to-no attention given to using them as classifiers. As far as we are aware, only \cite{ca_acc} report CA accuracies on common image datasets, they find that a CA achieves 64\% accuracy on CIFAR-10, down from a standard classifier's 93\% accuracy. \cite{ca_acc} do not explore the interpretabillity of their CA, as it was not the focus of their work.

\section{Methodology}
We begin by training a simple CA on the MNIST-10 dataset. We use a standard PixelCNN architecture \cite{pixelcnn}, and to condition on class we provide the class embedding as an auxiliary input, which is summed into output of every hidden layer. During training the model is trained to maximize the probability of the input with the correct class as auxiliary input. At test time we perform the generation process 10 times, with a different class as auxiliary input each time, and classify the input as belonging to the class which produced the highest probability of $x$.

Next, we train a standard ResNet classifier on the MNIST training set, and use it to label the test set by assigning test images to the class with the largest logit. We then train a CA on the MNIST training set as well as the test set with predicted labels. We refer to this model as conditional autoregressor with oracle (CA+O).

\section{Results}
\subsection{Classification Performance}
\begin{figure}
    \centering
    \includegraphics{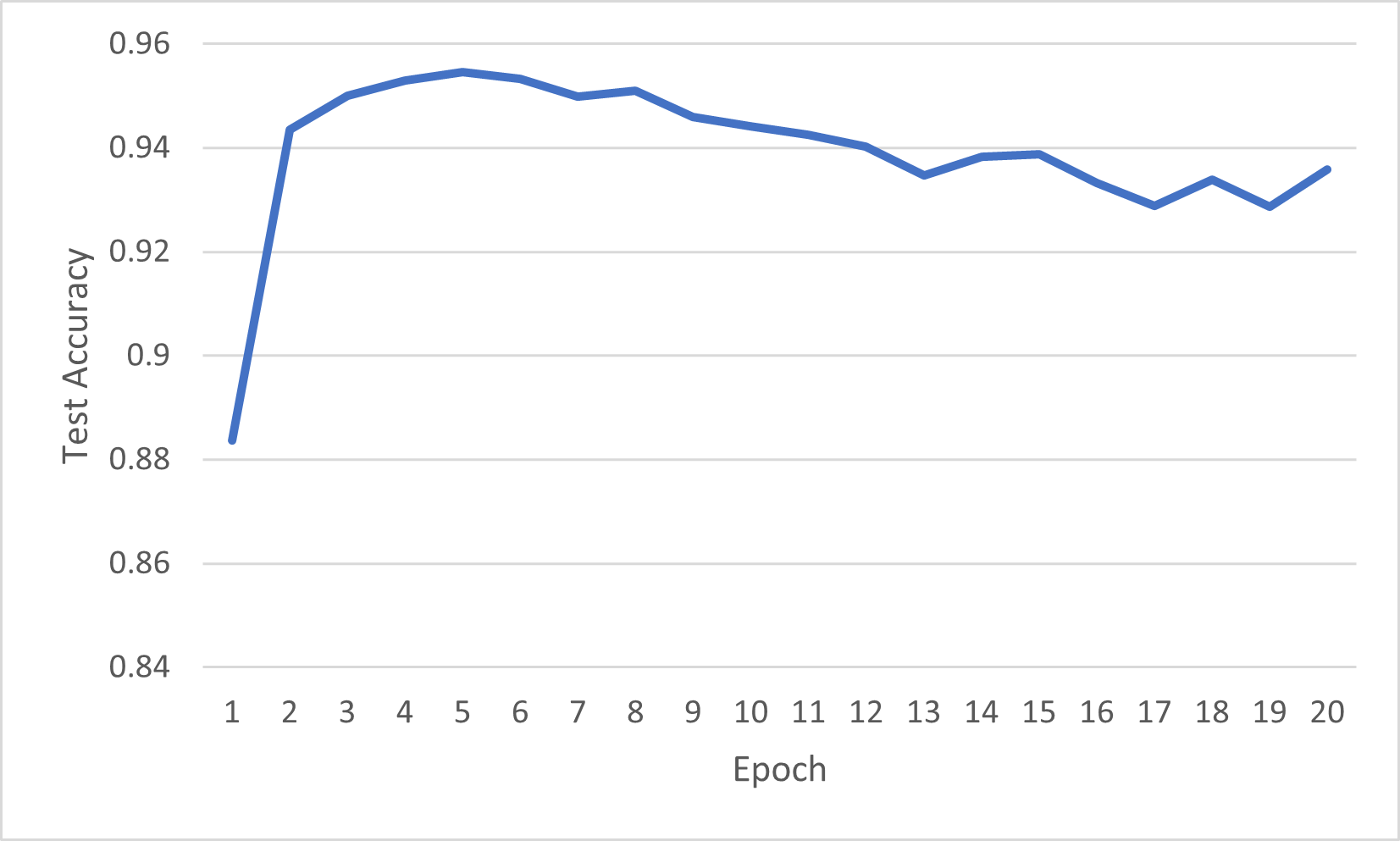}
    \caption{MNIST-10 test accuracy of a CA over time}
    \label{fig:CAacc}
\end{figure}

\begin{center}
\begin{tabular}{ |c||c|c|c| } 
 \hline
 Model & CA & ResNet & CA+O \\ 
 \hline
 Test Accuracy & 95.45 & 99.45 & 99.44 \\ 
 \hline
\end{tabular}
\end{center}

As seen in figure \ref{fig:CAacc}, after just 5 epochs of training, the CA begins to overfit and its test accuracy begins to decrease. The highest achieved accuracy is 95.45\%, significantly below that of the standard ResNet baseline.
However, we find that the CA+O, which is trained on the labels provided by the ResNet classifier, is eventually able to match the performance of the baseline. This suggests that CA are significantly more prone to overfitting than standard classifiers, however they do still have similar modelling capacity.

\subsection{Interpretabillity}
\begin{figure}[!ht]
\centerline{
    \subfloat{
    \includegraphics[width=.1\textwidth]{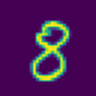}
    }
    \subfloat{
    \includegraphics[width=.1\textwidth]{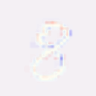}
    }
    \subfloat{
    \includegraphics[width=.1\textwidth]{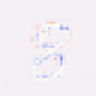}
    }
    \subfloat{
    \includegraphics[width=.1\textwidth]{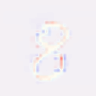}
    }
    \subfloat{
    \includegraphics[width=.1\textwidth]{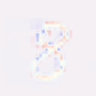}
    }
    \subfloat{
    \includegraphics[width=.1\textwidth]{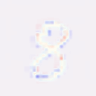}
    }
    \subfloat{
    \includegraphics[width=.1\textwidth]{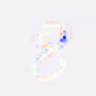}
    }
    \subfloat{
    \includegraphics[width=.1\textwidth]{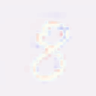}
    }
    \subfloat{
    \includegraphics[width=.1\textwidth]{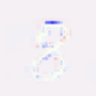}
    }
    \subfloat{
    \includegraphics[width=.1\textwidth]{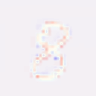}
    }
    \subfloat{
    \includegraphics[width=.1\textwidth]{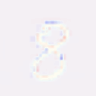}
    }
    }
    
    \centerline{
    \subfloat{
    \includegraphics[width=.1\textwidth]{input_9322}
    }
    \subfloat{
    \includegraphics[width=.1\textwidth]{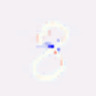}
    }
    \subfloat{
    \includegraphics[width=.1\textwidth]{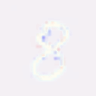}
    }
    \subfloat{
    \includegraphics[width=.1\textwidth]{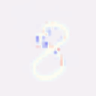}
    }
    \subfloat{
    \includegraphics[width=.1\textwidth]{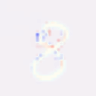}
    }
    \subfloat{
    \includegraphics[width=.1\textwidth]{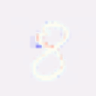}
    }
    \subfloat{
    \includegraphics[width=.1\textwidth]{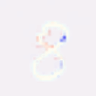}
    }
    \subfloat{
    \includegraphics[width=.1\textwidth]{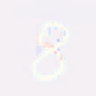}
    }
    \subfloat{
    \includegraphics[width=.1\textwidth]{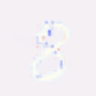}
    }
    \subfloat{
    \includegraphics[width=.1\textwidth]{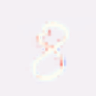}
    }
    \subfloat{
    \includegraphics[width=.1\textwidth]{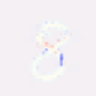}
    }
    }
    
    \centerline{
    \subfloat{
    \includegraphics[width=.1\textwidth]{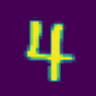}
    }
    \subfloat{
    \includegraphics[width=.1\textwidth]{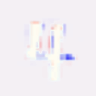}
    }
    \subfloat{
    \includegraphics[width=.1\textwidth]{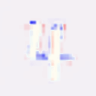}
    }
    \subfloat{
    \includegraphics[width=.1\textwidth]{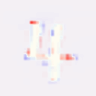}
    }
    \subfloat{
    \includegraphics[width=.1\textwidth]{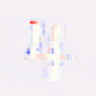}
    }
    \subfloat{
    \includegraphics[width=.1\textwidth]{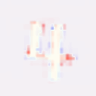}
    }
    \subfloat{
    \includegraphics[width=.1\textwidth]{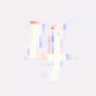}
    }
    \subfloat{
    \includegraphics[width=.1\textwidth]{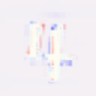}
    }
    \subfloat{
    \includegraphics[width=.1\textwidth]{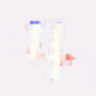}
    }
    \subfloat{
    \includegraphics[width=.1\textwidth]{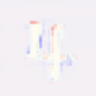}
    }
    \subfloat{
    \includegraphics[width=.1\textwidth]{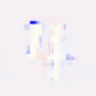}
    }
    }
    
    \centerline{
    \subfloat{
    \includegraphics[width=.1\textwidth]{input_8452}
    }
    \subfloat{
    \includegraphics[width=.1\textwidth]{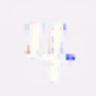}
    }
    \subfloat{
    \includegraphics[width=.1\textwidth]{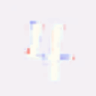}
    }
    \subfloat{
    \includegraphics[width=.1\textwidth]{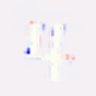}
    }
    \subfloat{
    \includegraphics[width=.1\textwidth]{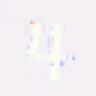}
    }
    \subfloat{
    \includegraphics[width=.1\textwidth]{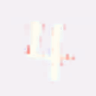}
    }
    \subfloat{
    \includegraphics[width=.1\textwidth]{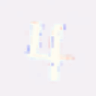}
    }
    \subfloat{
    \includegraphics[width=.1\textwidth]{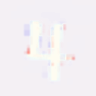}
    }
    \subfloat{
    \includegraphics[width=.1\textwidth]{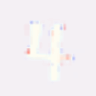}
    }
    \subfloat{
    \includegraphics[width=.1\textwidth]{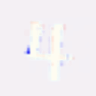}
    }
    \subfloat{
    \includegraphics[width=.1\textwidth]{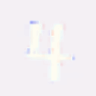}
    }
    }

    \caption{Heatmaps of $logp(x_i|x_{:i}, y)-logp(x_i|x_{:i})$ on a test 8 (rows 1 and 2) and a test 4(rows 3 and 4). Rows 1 and 3 are generated by CA, rows 2 and 4 are generated by CA+O.  Heatmaps shown are ordered by classes 0 (left) through 9 (right). Red indicates positive information, blue is negative.}
    \label{fig:heatmaps}
    
\end{figure}

We generate explanation heatmaps for the CA and CA+O models by plotting, for each pixel $i$, $logP(x_i|x_{:i},y) - logP(x_i|x_{:i})$, the novel information that pixel $i$ contains about $y$. As seen in figure \ref{fig:heatmaps}, the heatmaps generally agree with our intuition for which parts of the digit are informative. For example, the '8' image has strong negative information for class 3 on the left side of the top loop: the fact that these pixels are white is a strong indication that this image is not a 3.
We also find that the explanation heatmaps produced by the CA+O model are qualitatively the same as that of the CA model. Whilst there are some slight variations between them, they do tend to agree on what the significant parts of the image are. This suggests that, even though the CA+O model has been effectively trained on these images long past the point of overfitting, its explanation heatmaps are still valid.

\section{Discussion}
Our findings show that CA are a promising candidate for building interpretable machine learning models on high dimensional data. Training a simpler surrogate model to match the predictions of a black-box classifier is a widely used technique to aid interpretabillity. This approach can be seen in knowledge distillation based approaches and in LIME/SHAP. Existing methods use simple models, such as a decision tree or linear classifier. These low capacity models are often unable to match the performance of the black-box classifier on high-dimensional data, even when the surrogates are trained directly on the black-box predictions. Our work suggests that it may not be necessary to restrict surrogates to low capacity models, as even deep neural network CA are still interpretable.

Additionally, if CA models could be improved so that they can generalize reliably, then one could use a CA as a high performance and interpretable model without the need to train with a standard classifier oracle.

\bibliographystyle{plain}
\bibliography{refs.bib}

\end{document}